\documentclass{bmvc2k}

\usepackage{graphicx}
\usepackage{comment}
\usepackage{amsmath,amssymb} 
\usepackage{color}
\usepackage{booktabs}
\usepackage{multirow}
\usepackage{array}
\usepackage{microtype}
\usepackage{bbding}
\usepackage{enumitem}
\definecolor{limegreen}{HTML}{32CD32}
\newcommand{\ours}[0]{SketchDreamer}
\newcommand{\sketch}{\mathcal{S}}

\newcommand{\renderer}{\mathcal{R}}
\newcommand{\initsketch}{\mathcal{S^*}}

\newcommand{\x}[0]{\mathbf{x}}
\newcommand{\z}[0]{\mathbf{z}}

\graphicspath{{./figures/}}

\title{SketchDreamer: Interactive Text-Augmented Creative Sketch Ideation}

\addauthor{Zhiyu Qu}{z.qu@surrey.ac.uk}{1,2}
\addauthor{Tao Xiang}{t.xiang@surrey.ac.uk}{1,2}
\addauthor{Yi-Zhe Song}{y.song@surrey.ac.uk}{1,2}

\addinstitution{
 SketchX, CVSSP, \\ University of Surrey
}
\addinstitution{
 iFlyTek-Surrey Joint Research Centre \\ on Artificial Intelligence
}

\runninghead{Z.Qu, T.Xiang, Y.Song}{Sketch Dreamer}

\def\ie{\emph{i.e}\bmvaOneDot}
\def\eg{\emph{e.g}\bmvaOneDot}

\begin{document}

\maketitle

\begin{abstract}
Artificial Intelligence Generated Content (AIGC) has shown remarkable progress in generating realistic images. However, in this paper, we take a step ``backward'' and address AIGC for the most rudimentary visual modality of human sketches. Our objective is on the creative nature of sketches, and that creative sketching should take the form of an interactive process. We further enable text to drive the sketch ideation process, allowing creativity to be freely defined, while simultaneously tackling the challenge of ``I can't sketch''. We present a method to generate controlled sketches using a text-conditioned diffusion model trained on pixel representations of images. Our proposed approach, referred to as \ours{}, integrates a differentiable rasteriser of B\'ezier curves that optimises an initial input to distil abstract semantic knowledge from a pretrained diffusion model. We utilise Score Distillation Sampling to learn a sketch that aligns with a given caption, which importantly enable both text and sketch to interact with the ideation process. Our objective is to empower non-professional users to create sketches and, through a series of optimisation processes, transform a narrative into a storyboard by expanding the text prompt while making minor adjustments to the sketch input. Through this work, we hope to aspire the way we create visual content, democratise the creative process, and inspire further research in enhancing human creativity in AIGC. The code is available at \url{https://github.com/WinKawaks/SketchDreamer}.
\end{abstract}

\section{Introduction}  \label{sec:intro}

Artificial Intelligence Generated Content (AIGC) has been making tremendous progress~\cite{imagen, dalle2, rombach2022high} in generating high-quality images that are often indistinguishable from real photographs. However, despite this significant advancement, there remains a critical lack of creativity embedded in the AIGC process~\cite{li2022mat, wang2022pretraining}. In this paper, we aim to tackle this issue by exploring the most fundamental form of visual communication known to humanity since prehistoric times - human sketches~\cite{sketchrnn,das2020beziersketch,ge2020creative,pang2020solving,das2021sketchode,yang2022finding}. Sketches are an essential means of conveying ideas, emotions, and information, and their creative nature makes them an ideal candidate for exploring new avenues in AIGC~\cite{voynov2022sketch,mikaeili2022sked,zhang2023adding}.

\begin{figure*}[t]
\centering
    \begin{subfigure}[h]{\textwidth}
        \includegraphics[width=\textwidth]{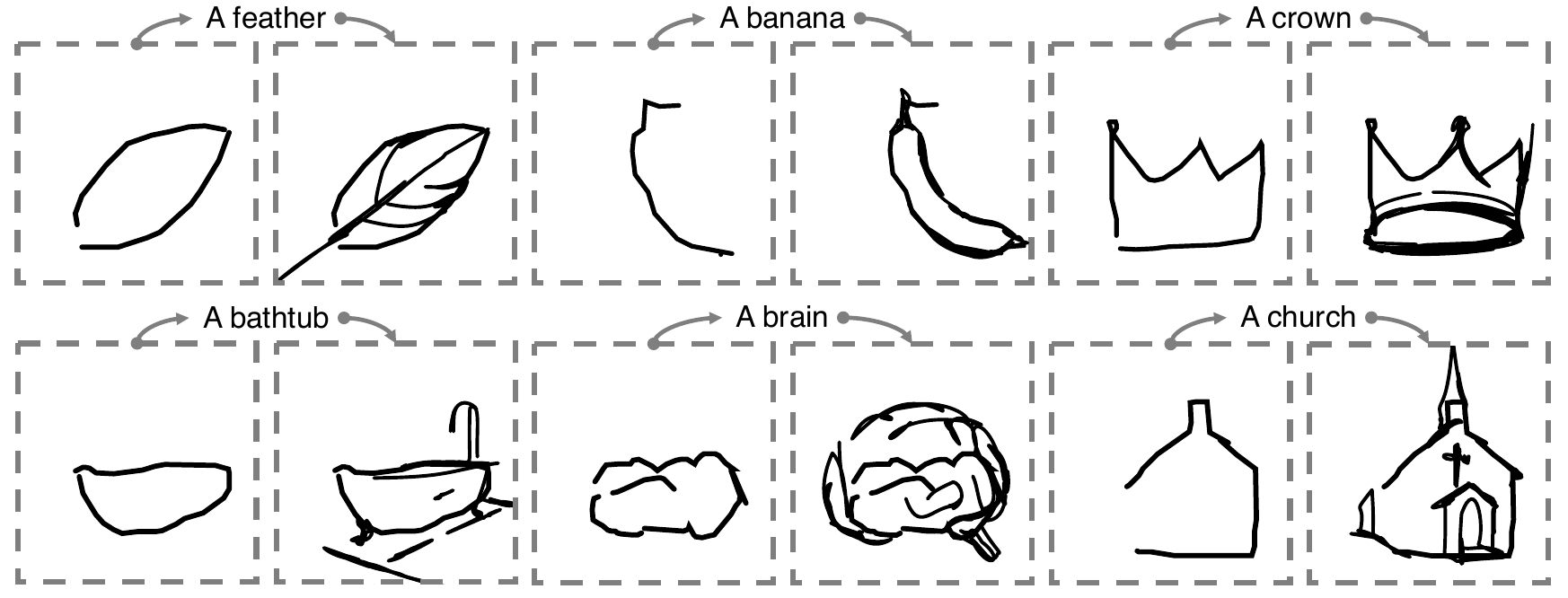}
        \vspace{-1.8em}
        \caption{Generation by single round.}
        \vspace{2mm}
    \end{subfigure}%
    
    \begin{subfigure}[h]{\textwidth}
        \includegraphics[width=\textwidth]{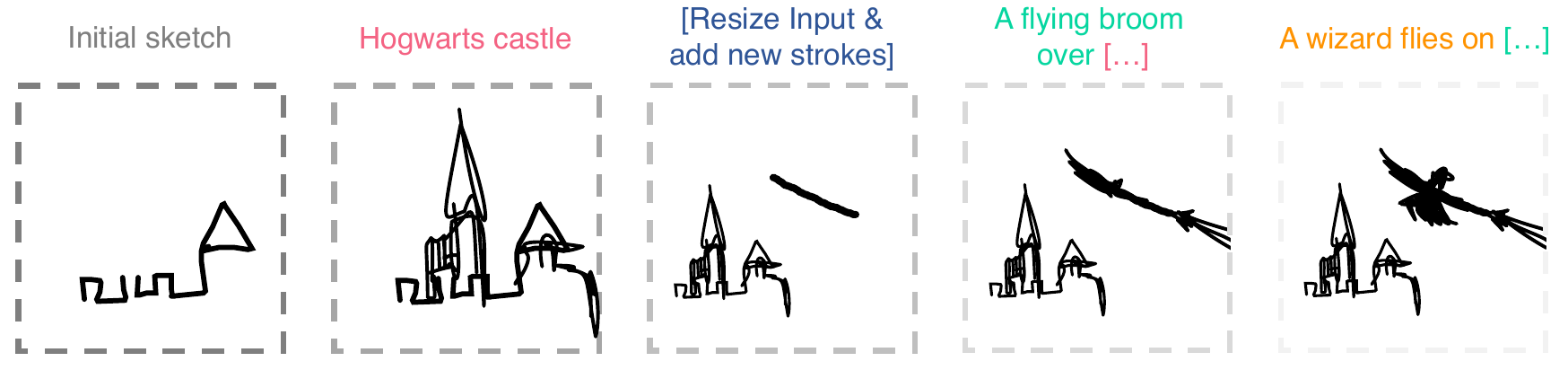}
        \vspace{-1.8em}
        \caption{Interactive generation by multiple rounds.}
        \vspace{2mm}
    \end{subfigure}
    
\caption{(a) Given an initial sketch and a textual prompt, \ours{} can generate a new sketch that closely corresponds to the text and initial sketch conditions. (b) By iteratively expanding the text prompt (\texttt{[…]} indicating the previous text prompt) and making minor adjustments such as resizing, relocating or adding new strokes, our model can easily generate a storyboard interactively. It is important to note that all initial sketches used in this paper were selected from QuickDraw~\cite{sketchrnn}, which accurately reflects the average sketching ability of non-professional users.}
\label{fig:overview}
\vspace{-1em}
\end{figure*}

Existing AIGC approaches, such as DALL-E 2 \cite{dalle2}, Imagen \cite{imagen}, and Latent Diffusion \cite{rombach2022high}, which mainly focus on high-quality image generation, have been limited in their ability to capture human creativity in the content creation process. Most recent works in AIGC use a multi-stage approach that allows user edits~\cite{li2022mat,yang2022diffusion,lugmayr2022repaint,saharia2022palette,wang2022zero,zhang2023dreamface}, but the newly generated content is essentially a new image. However, with sketches, we believe it is possible to capture a truly interactive and procedural creative visual content creation process, where new strokes are added to the previous drawn sketch, instead of producing a new sketch from scratch. This approach mimics the way humans draw and encourages a more interactive and dynamic creative process.

Furthermore, our approach uniquely allows for both text and sketch to inform the creative ideation process, which is crucial in maximally injecting creativity into the sketching process. This is because sketching and text ideation are not mutually exclusive, and our method allows for the two modalities to interact and influence each other in the ideation process. Importantly, this also addresses the ``I can't sketch'' problem (see Fig.~\ref{fig:overview} (a)), thereby democratising the ideation process and enabling novice users to create sketches effortlessly. By allowing for a more interactive and fluid ideation process, we hope to pave the way for a more inclusive and democratised approach to visual content creation, where creativity can be expressed more freely and inclusively (see Fig.~\ref{fig:overview} (b)).

Closest to our setup would be the very recent work~\cite{jain2022vectorfusion} on optimising SVG paths using a differentiable rasteriser to generate SVG images aligned with captions. Their reliance on text-only conditional diffusion models~\cite{rombach2022high} however largely limits the creative process and more importantly, it fundamentally lacks the concept of interactive ideation. To address these limitations, we introduce \ours{}, a method that generates controllable sketches from text captions based on initial sketches, while freely allowing for consequent ideation prompts via both text and sketch. The effectiveness of our method is demonstrated in Fig.~\ref{fig:overview}.

Following previous works~\cite{Li:2020:DVG,CLIPDraw,styleclipdraw,vinker2022clipasso,jain2022vectorfusion}, we utilise a differentiable B\'ezier curves renderer to allow for a flexible sketch representation. We further use a score distillation sampling (SDS) loss~\cite{poole2022dreamfusion} to enhance control over sketch generation while preserving coherence with the caption. To empower users to further engage in the creative process, our approach integrates ControlNet~\cite{zhang2023adding}, which supports multiple types of extra conditions (\eg, canny edges, HED boundaries, user scribbles, human poses, semantic maps, depths, \textit{etc}.) to manipulate the diffusion process instead of a text-only conditional diffusion model. The initial sketch serves a dual purpose: it acts as both the default image for the renderer and the scribble condition for ControlNet~\cite{zhang2023adding}, providing a more controllable and interactive creative process.

Our contributions are threefold: (i) we introduce the problem of creative sketch ideation in the context of recent approaches in AIGC that lack creativity, (ii) we propose a novel paradigm for interactive sketch generation that enables users to effortlessly create sketches via both text and sketch prompts, and (iii) we demonstrate the effectiveness of our approach through qualitative results and a human study that validates our method.

\begin{figure*}[!ht]
    \centering
    \includegraphics[width=\linewidth]{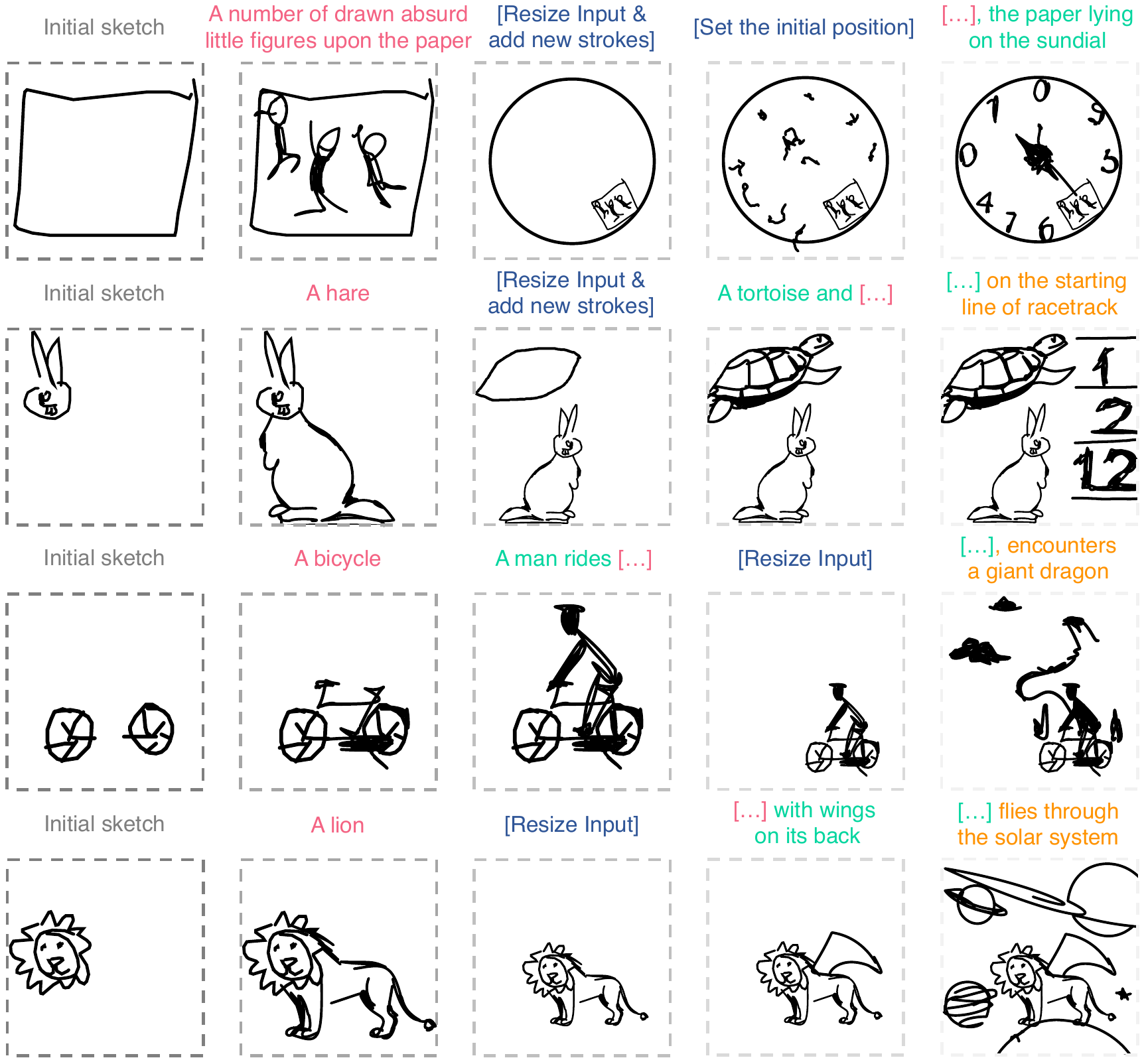}
    \caption{Various storyboards created by \ours{} are shown, with \texttt{[…]} indicating the previous text prompt. The initial sketches of the first column are all selected from QuickDraw~\cite{sketchrnn}, which reflects the average sketch ability of non-professional users. Our model provides multiple adjustable settings, such as resizing, relocating, adding new strokes, or setting the initial position manually, allowing users to generate their own unique storyboard by continually expanding the text prompt. The prompts are based on \textit{The Adventure of the Dancing Men} from \textit{The Canon of Sherlock Holmes}, \textit{The Tortoise and the Hare} from \textit{Aesop's Fables} and our own imagination, respectively.}
    \label{fig:story}
    \vspace{-1em}
\end{figure*}

\vspace{-1em}
\section{Related Work}  \label{sec:related_work}

\paragraph{Diffusion models.}
Diffusion models~\cite{ddpm,ddim,song2020score} have gained considerable attention due to their stability, diversity, and scalability. Due to these advantages, diffusion models have found applications in diverse fields, including image translation~\cite{rombach2022high,su2022dual,seo2022midms}, image editing~\cite{meng2021sdedit, Kim_2022_CVPR}, and conditional generation~\cite{kanizo2013palette,rombach2022high,zhang2023adding}. Particularly, text-to-image generation has been emphasised, and various guidance techniques~\cite{classifierfree,dhariwal2021diffusion,hong2022improving} have been introduced to improve it. \cite{Kim_2022_CVPR,Nichol2022GLIDETP} utilises CLIP~\cite{clip} guidance to enable text-to-image generation, followed by large-scale text-to-image diffusion models~\cite{imagen,dalle2,rombach2022high}.
These models' emergence has led to the extensive utilisation of pretrained text-to-image models for tasks such as adding additional conditions~\cite{wang2022pretraining,zhang2023adding} or performing manipulations~\cite{gal2022image,kwon2022diffusion,ruiz2022dreambooth}. In our work, we use ControlNet~\cite{zhang2023adding} instead of vanilla diffusion models for better constraint of sketch generation.

\vspace{-1em}

\paragraph{Sketch generation.}
Previous studies primarily focused on the generation of sketches from images using an image-to-image translation approach~\cite{isola2017image,song2018learning,li2019im2pencil}. However, facilitated by the availability of large sketch datasets with stroke-level information~\cite{sketchrnn}, researchers have become increasingly interested in methods that can generate human-like sketches without images. One notable early work is SketchRNN~\cite{sketchrnn}, which employs a sequence-to-sequence variational autoencoder to model the temporal sequence of vector coordinates in a sketch. Subsequent approaches~\cite{chen2017sketch, cao2019ai} incorporated convolutional encoders to model the spatial information of sketches.
In addition to generating complete sketches, there is also been a focus on sketch completion approaches, where missing parts are generated based on a partial sketch. SketchGAN~\cite{liu2019sketchgan} utilises a conditional Generative Adversarial Network (GAN) model to generate the missing part. Inspired by language pretraining approaches~\cite{bert}, Sketch-BERT~\cite{lin2020sketch} names the task of completing the missing part of sketches as ``sketch gestalt''. DoodlerGAN~\cite{ge2020creative} aims to generate creative sketches by combination of novel compositions of part appearances from two creative datasets.

In recent years, the rapid development of visual-language models~\cite{clip} has inspired several studies to investigate the utilisation of pretrained vision-language models for guiding sketch generation. CLIPDraw~\cite{CLIPDraw} utilises CLIP's image-text cosine similarity loss~\cite{clip} to generate vector graphics from text prompts, employing a procedure similar to \cite{mordvintsev2018differentiable, goh2021multimodal}. StyleCLIPDraw~\cite{styleclipdraw} incorporates conditioning on images with an auxiliary style loss. CliPasso~\cite{vinker2022clipasso} introduces a new geometric loss to convert an image to a sketch with various levels of abstraction. Recently, some work~\cite{jain2022vectorfusion, iluz2023word} has shifted the perspective from CLIP to diffusion models. VectorFusion~\cite{jain2022vectorfusion} is the most relevant work in our context, but it is purely text-driven. In contrast, our method generates high-quality sketches under control, conditioned on both text and initial sketches.

\vspace{-1em}

\paragraph{Vector graphics.} Extensive research exists on stroke-based rendering, contour visualisation, and feature line rendering, which could be comprehensively reviewed in surveys~\cite{Hertzmann2003-survey, Bnard2019LineDF}.
Vector representations find extensive usage in various sketch-related tasks and applications, leveraging multiple deep learning models including RNNs~\cite{sketchrnn}, CNNs~\cite{li2018sketch}, GNNs~\cite{yang2021sketchgnn}, Transformers~\cite{ribeiro2020sketchformer, lin2020sketch, qu2023sketchxai}, GANs~\cite{Varshaneya2021TeachingGT} and reinforcement learning algorithms~\cite{Zhou2018LearningTS, mellor2019unsupervised, Ganin2018SynthesizingPF}.
Recent advancements of differentiable rendering algorithms~\cite{Zheng2019StrokeNetAN, Mihai2021DifferentiableDA, Li:2020:DVG} enable the manipulation and synthesis vector content through raster-based loss functions. In our work, we utilise a differentiable rasteriser~\cite{Li:2020:DVG}, which is capable of processing various types of strokes and curves, including B\'ezier curves and computing the gradient of the rendered image with respect to the parameters of the these primitives.

\vspace{-1em}
\section{Methodology}  \label{sec:methodology}

\subsection{Preliminaries}\label{sec:preliminaries}

\paragraph{Diffusion models.} 
Diffusion models belong to a flexible class of likelihood-based generative models that learn a distribution through denoising. During the training process, diffusion models optimise a variational bound on the likelihood of real data samples~\cite{pmlr-v37-sohl-dickstein15}, following a similar approach to variational autoencoders~\cite{Kingma2014AutoEncodingVB}. This bound can be expressed as a weighted combination of denoising objectives~\cite{ddpm} as shown in Eq.~\ref{eq:ddpm}:
\begin{equation} \label{eq:ddpm}
\mathcal{L}_\text{DDPM}(\phi, \x) = \mathbb{E}_{t, \epsilon} \left[ w(t) \| \epsilon_\phi(\alpha_t \x + \sigma_t \epsilon) - \epsilon \right \|^2_2],
\end{equation}
where $\x$ represents a real data sample, $w(t)$ is the weighting function and $t\in \{1, 2, \ldots T\}$ is a uniformly sampled timestep scalar that indexes noise schedules $\alpha_t, \sigma_t$~\cite{kingma2021on}. The noise term $\epsilon$ has the same dimension as the image sampled from the known Gaussian prior. Noise is added by interpolation to preserve variance. For images, $\epsilon_\phi$ is a learned denoising autoencoder and commonly implemented as a U-Net~\cite{unet, ddpm} that predicts the noise content of its input.

In the context of text-to-image generation, the U-Net architecture is often conditioned on a caption $c$, resulting in $\epsilon_\phi(\x, c)$. This condition is typically achieved through cross-attention layers and text features of a language model~\cite{imagen,dalle2,rombach2022high}. However, conditional diffusion models may generate outputs that lack coherence with the provided captions. In order to enhance the relevance of the caption, \cite{classifierfree} superconditions the model by scaling up conditional model outputs and deviating from a generic unconditional prior that disregards the caption $c$:

\begin{equation}
\hat{\epsilon}_\phi(\x, c) = (1 + \omega) * \epsilon_\phi(\x, c) - \omega * \epsilon_\phi(\x).
\end{equation}

\paragraph{Score Distillation Sampling.} 
DreamFusion~\cite{poole2022dreamfusion} proposed a novel approach that utilises a pretrained text-to-image diffusion model as a loss function. Their method introduces SDS loss, which enables the assessment of the similarity between an image $\x$ and a corresponding caption $c$:
\begin{align}
    \mathcal{L}_\text{SDS} &= \mathbb{E}_{t, \epsilon} \left[ \sigma_t / \alpha_t w(t) \text{KL}(q(\x_t|g(\theta); c, t) \| p_\phi(\x_t ; c,t ))\right],
\end{align}
where $p_\phi$ represents the distribution learned by the frozen diffusion model. $q$ corresponds to a unimodal Gaussian distribution centred around a learned mean image $g(\theta)$. 
The SDS loss treats the process of sampling as an optimisation problem, allowing the optimisation of an image or a differentiable image parameterisation (DIP)~\cite{mordvintsev2018differentiable} with respect to $\mathcal{L}_\text{SDS}$ to align it with the conditional distribution of the teacher model. It draws inspiration from probability density distillation~\cite{Oord2018ParallelWF}. Importantly, SDS solely requires access to a pixel-space prior $p_\phi$ parameterised by the denoising autoencoder $\hat{\epsilon}_\phi$, and it does not rely on a prior over the parameter space $\theta$. In practice, SDS computes the difference of the added noise and predicted noise as per-pixel gradient:
\begin{equation} \label{eq:sds}
    \nabla_\theta \mathcal{L}_\text{SDS} = \mathbb{E}_{t, \epsilon}\left[w(t)\left(\hat\epsilon_\phi(\x_t; c, t)  - \epsilon\right) {\frac{\partial \x}{\partial \theta}}\right].
\end{equation}

\begin{figure*} [!t]
\centering
\includegraphics[width=\linewidth]{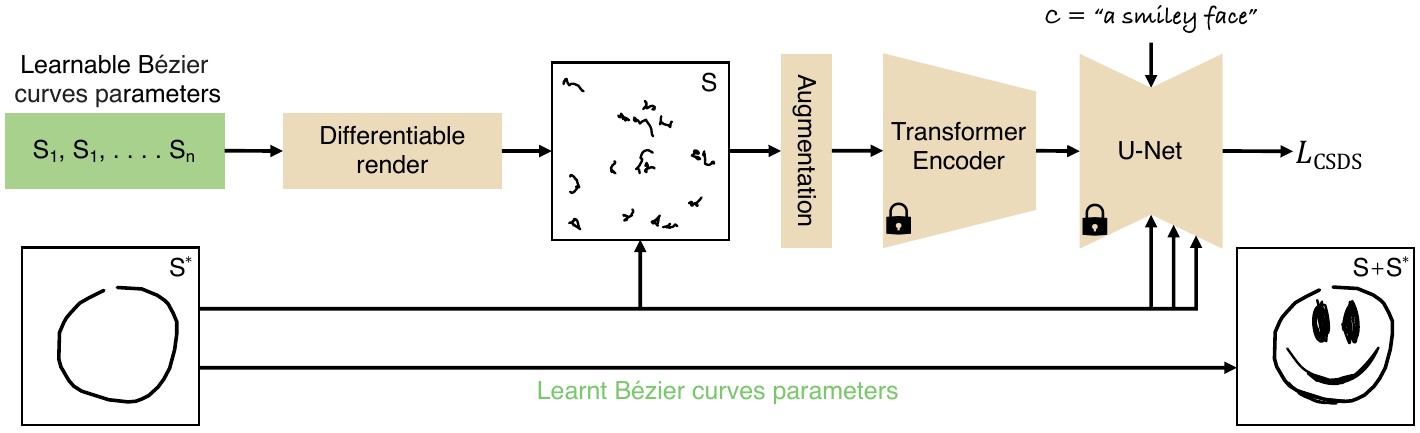}
\caption{
An overview of the training procedure for \ours{} is as follows: We start with an initial sketch $\initsketch$ and the number of strokes $n$. Using a differentiable rasteriser, $\renderer$, we create a rasterised sketch $\sketch$. Next, we feed $\sketch + \initsketch$ into a pretrained diffusion model while setting $\initsketch$ as an additional input condition. We apply data augmentations, encode into a latent space, compute the Score Distillation Sampling loss~\cite{poole2022dreamfusion} on the latent, and backpropagate through the above modules to update the parameters of B\'ezier curves.
}
\label{fig:parameters}
\vspace{-1em}
\end{figure*}

\vspace{-1em}
\subsection{\ours{}}\label{sec:architecture}

We initialise a sketch as a set of $n$ strokes $\{s_1, .. s_n\}$ placed on a white background. Each stroke is represented by a two-dimensional B\'ezier curve with four fixed control points $s_i = \{p_i^j\}_{j=1}^4 = \{(x_i,y_i)^j\}_{j=1}^4$. For simplicity the optimisation process, we optimise only the position of control points, while fixing other attributes of strokes, including width, colour and opacity. Moreover, we require a fixed initial sketch $\initsketch$ which can be either some simple strokes or any subplot of a storyboard. We feed the parameters of the strokes to a differentiable rasteriser $\renderer$, which produces a rasterised sketch $\sketch = \renderer(\{p_1^j\}_{j=1}^4, ... \{p_n^j\}_{j=1}^4) = \renderer(s_1, .. s_n)$.

The training procedure of our method is shown in Fig.~\ref{fig:parameters}.
Given a text condition $c$, our goal is to synthesise the corresponding sketch $\sketch$ so that $\sketch + \initsketch$ matches $c$ best. To integrate the existing latent diffusion models (LDM) with human sketches, it is necessary to constrain the diversity and manipulate diffusion model generation processes. Therefore, we use a sketch-conditional diffusion model (\ie, ControlNet~\cite{zhang2023adding}) instead of vanilla LDM and set the initial sketch $\initsketch$ as the extra condition in addition to the text condition $c$.

We start with the randomly initial locations of the strokes. Next, in each step of the optimisation we feed the stroke parameters to a differentiable rasteriser $\renderer$ to produce the rasterised sketch. Like \cite{CLIPDraw}, we augment the resulting sketch $\sketch$, as well as the initial sketch $\initsketch$ with perspective transform and random crop to get a $512\times 512$ sketch $\sketch_\text{aug}$. Then, we feed the augmented sketches into the LDM model to compute the SDS loss in latent space using the LDM encoder $E_\phi$, predicting $\z = E_\phi(\sketch_\text{aug})$.
For each iteration of optimisation, we diffuse the latents with random noise $\z_t = \alpha_t \z + \sigma_t \epsilon$, denoise with the teacher model $\hat{\epsilon}_\phi(\mathbf{z}_t; c, t, \initsketch)$, and optimise the SDS loss using a latent-space modification of Equation~\ref{eq:sds}:

\begin{equation}
    \nabla_\theta \mathcal{L}_\text{CSDS} = \\\quad\mathbb{E}_{t, \epsilon} \left[ w(t) \Big(\hat{\epsilon}_\phi(\alpha_t \z_t + \sigma_t \epsilon; c, t, \initsketch)  - \epsilon\Big) \frac{\partial \z}{\partial \sketch_\text{aug}} \frac{\partial \sketch_\text{aug}}{\partial \theta} \right]\\.
\end{equation}

During backpropagation process, the term $\partial \sketch_\text{aug} / \partial \theta$ is computed with auto-differentiation through the augmentations and differentiable rasteriser. $\mathcal{L}_\text{CSDS}$ is an adaptation of $\mathcal{L}_\text{SDS}$ with extra conditions, which treats the rasteriser, data augmentation and frozen LDM encoder as a single image generator with optimisable parameters $\theta$ for the B\'ezier curves.

\vspace{-1em}
\section{Experiments}  \label{sec:experiments}

\subsection{Experimental settings}

Following the implementation of \cite{jain2022vectorfusion}, we initialise B\'ezier curves with 5 segments, a fixed stroke width and a fixed black colour. We apply random affine augmentations to the sketch before passing them as inputs to the diffusion model. Specifically, we use RandomPerspective with a probability of 0.7 and a distortion scale of 0.2, and RandomResizedCrop, which resizes the sketches from 600$\times$600 to 512$\times$512. These augmentations improve the quality of the generated sketch and make the optimisation process less susceptible to adversarial samples~\cite{CLIPDraw,vinker2022clipasso,jain2022vectorfusion}. We optimise the sketches for 1000 iterations using the Adam optimiser~\cite{adam2014kingma}, with a learning rate set to 1. In our setting, we use a guidance scale of $\omega$=100.

We conduct a comparison of our method with two different settings of VectorFusion, namely VectorFusion (VectorFusion with text only) and VectorFusion Init (VectorFusion with text and an initial sketch as input to LDM). The qualitative and quantitative evaluations are presented as follows.

\begin{figure*}[!ht]
    \centering
    \includegraphics[width=\linewidth]{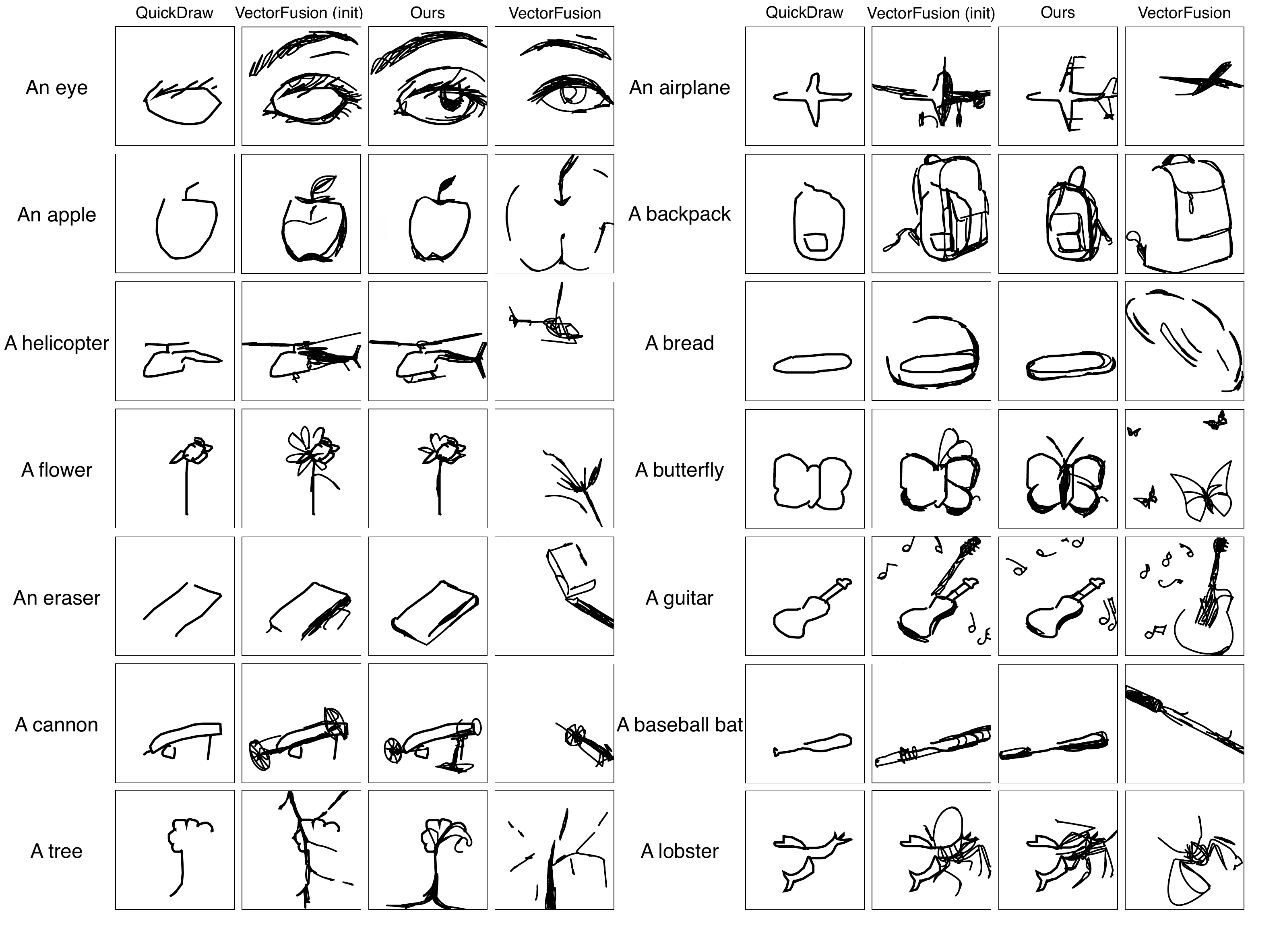}
    \vspace{-6mm}
    \caption{Comparison to existing text-to-sketch works. The first two columns show the labels and initial sketches selected from QuickDraw. We use a special prompt modifiers to encourage an appropriate style for sketch generation of single objects: ...on a white background. Taking into account a suboptimal initial sketch, our approach is capable of generating results that retain fidelity to the initial sketch while exhibiting a degree of artistic flair.}
    \label{fig:exp_comparison}
    \vspace{-1em}
\end{figure*}

\begin{figure*}[!h]
    \centering
    \includegraphics[width=\linewidth]{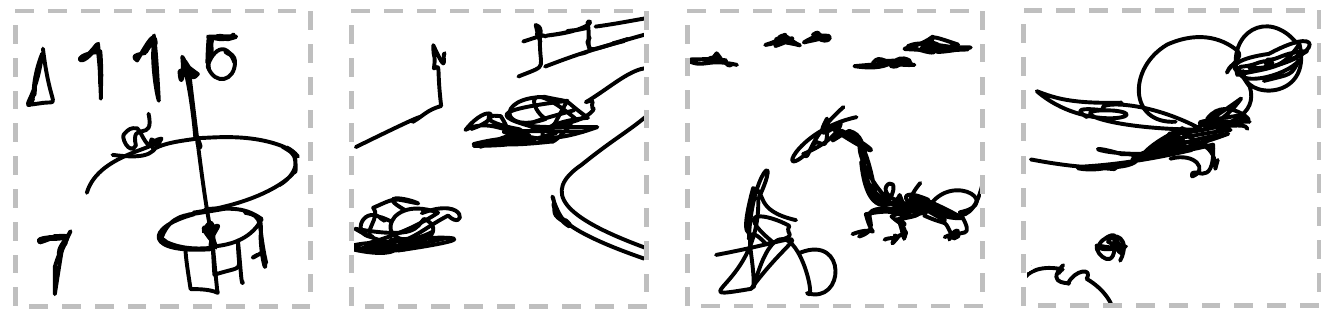}
    \vspace{-6mm}
    \caption{As VectorFusion~\cite{jain2022vectorfusion} is not capable of supporting progressive interaction, we directly utilise the 4 completed captions in Fig.~\ref{fig:story} to generate sketches via VectorFusion. The captions are ``A number of drawn absurd little figures upon the paper, the paper lying on the sundial'', ``A tortoise and a hare on the starting line of racetrack'', ``A man rides a bicycle, encounters a giant dragon'' and ``A lion with wings on its back flies through the solar system'', respectively.}
    \label{fig:evaluatio_story}
    \vspace{-1em}
\end{figure*}

\subsection{Qualitative evaluation}
\label{sec:experiments:qualitative}

Fig.~\ref{fig:exp_comparison} presents a qualitative comparison of \ours{} with other text-to-sketch synthesis methods, where all initialised sketches are obtained from QuickDraw\cite{sketchrnn}. QuickDraw contains 345 object categories with 75K sketches per category. During acquisition, the participants were given only 20 seconds to sketch an object. As a result, sketches of QuickDraw are more iconic and abstract, and some are not even finished. By inputting these simple sketches as the initial sketches into ControlNet, we effectively constrain the diffusion processes. Leveraging the powerful generation capabilities of LDM, it is possible to meticulously complete the sketches with intricate details. Compared to the VectorFusion and VectorFusion Init, our proposed method \ours{} produces significantly more controllable generation based on the initial sketches.

In Fig.~\ref{fig:story}, we show the progressive interaction of our work. As VectorFusion~\cite{jain2022vectorfusion} is not capable of supporting progressive interaction, we directly utilise the 4 completed captions to generate sketches via VectorFusion in Fig.~\ref{fig:evaluatio_story}. Comparing these sketches with Fig.~\ref{fig:story}, it is difficult for VectorFusion to have any control over the sketch appearance. In addition, we find that when the captions contain multiple objects, the sketches generated by VectorFusion may exhibit issues such as missing objects and poorer quality.

\subsection{Quantitative evaluation}
\label{sec:main_experiments}

It can be challenging to evaluate text-to-sketch synthesis due to the absence of target or ground truth sketches that can serve as references. To overcome this challenge, we construct a diverse evaluation dataset consisting of 128 captions obtained from prior studies and benchmarks in the field of text-to-image generation. The coherence between text and sketches is assessed using automated CLIP metrics. Similar to previous works, we employ both CLIP R-Precision and cosine similarity as evaluation metrics. Additionally, we conduct a user study to provide a more human-friendly metric for assessing the quality of the generated sketches. 

\textbf{CLIP Similarity and R-Precision.} The average cosine similarity of CLIP embeddings is computed between the generated images and their corresponding text captions, excluding any prompt engineering from the reference text. Higher CLIP Similarity scores suggest a higher degree of consistency between the text-image pairs. To provide a more interpretable metric, we calculate CLIP Retrieval Precision as proposed in~\cite{Park2021BenchmarkFC}. The R-Precision metric represents the percentage of sketches that achieve the highest CLIP Similarity with the correct input caption, out of the total 128 captions in our dataset.

\textbf{User study.} We conduct a user study with 50 participants to determine which method produces results that are most faithful to the initial sketches. In the user study, participants are presented with groups of sketches. Each group contains one initial sketch and three generated sketches, and participants are encouraged to consider both the fidelity to the initial sketch and the presence of creative elements when making their choices.

The results are presented in Tab.~\ref{tab:main_results}. Our initial sketches obtained from QuickDraw~\cite{sketchrnn} have a low quality, as evidenced by the R-Prec score of 28.91\%. Therefore, both our controllable and uncontrollable methods produced sketches that scored much higher than QuickDraw. The low score of the uncontrollable VectorFusion can be attributed to the fact that the model tends to generate sketches that fill the entire canvas, resulting in partially missing sketches if the positions of the B\'ezier curves are not well-suited to the context, as seen in the example of \texttt{[apple]} in Fig.~\ref{fig:exp_comparison}.

In Tab.~\ref{tab:main_results}, our method does not achieve the highest R-Prec and Sim scores. This is due to the poor quality of the initial sketch obtained from QuickDraw, which may not be suitable as a control condition when fed directly into LDM. As a result, VectorFusion Init may produce sketches similar to uncontrollable generation instead of controllable generation (see \texttt{[backpack]} in Fig.~\ref{fig:exp_comparison}). The ideal evaluation metric would be one that measures the similarity between the generated sketches and the initial sketches. Unfortunately, none of the available evaluation metrics~\cite{ssim, huynh2008scope, zhang2018perceptual} for measuring image similarity can be applied directly to sketches. To address this limitation, we conduct a user study. As shown in the right column of Table~\ref{tab:main_results}, the sketches generated by \ours{} are significantly more consistent with the initial sketches than those of the other methods.

\vspace{5mm}

\begin{table}[t]
  \centering
  \resizebox{0.7\linewidth}{!}{%
  \begin{tabular}{@{}l|c|cc|c@{}}
    \toprule
    Method & Control & R-Prec $\uparrow$ & Sim $\uparrow$ & User study $\uparrow$ \\ 
    \midrule
    VectorFusion & {\color{red}\XSolidBrush} & 57.03 & 30.91 & 8.75\% \\
    VectorFusion (init) & {\color{limegreen}\Checkmark} & \textbf{74.22} & \textbf{31.71} & 22.91\% \\
    \ours{} & {\color{limegreen}\Checkmark} & 72.94 & 31.50 & \textbf{68.34\%} \\
    \midrule
    QuickDraw & - & 28.91 & 29.33 & - \\
    \bottomrule
  \end{tabular}}
  \vspace{3mm}
  \caption{Evaluation of the consistency of text-to-sketch generations using 16 B\'ezier curves with input captions. Consistency is measured with CLIP R-Precision and CLIP similarity score ($\times$100). In the conducted user study, we randomly select a total of 20 sets of results, and the participants are requested to select the sketch that most closely resembled the initial sketch from each set.}
  \label{tab:main_results}
  \vspace{-2em}
\end{table}

\vspace{-1em}
\section{Limitations and Discussion}  \label{sec:limitations}

Our proposed method, \ours{}, has certain limitations. The performance of our proposed method can be influenced by dataset biases and quality issues, stemming from the use of the LDM algorithm~\cite{laionbias}. Nevertheless, we anticipate that the performance of our proposed method will enhance with the advancements in text-to-image models.

Another aspect worth noting is that \cite{vinker2022clipasso} emphasises the optimisation process is susceptible to the initialisation of B\'ezier curves. This susceptibility arises due to the challenge of finding a locally optimal solution in a highly non-convex function. They propose placing the initial strokes based on the salient regions of the target image to improve convergence towards semantic depictions, as the saliency map can provide useful prior information. However, we find it challenging to use text to initialise B\'ezier curves effectively. We technically adapt the approach of periodically removing paths with fill-colour opacity or area below a threshold and randomly reinitialising new curves, as proposed by~\cite{jain2022vectorfusion}. However, it is noteworthy to emphasise that the decision on where to initialise the strokes should be left to the users in future applications.

\vspace{-1em}
\section{Conclusion}  \label{sec:conclusion}

In conclusion, our proposed approach offers a new and innovative way of exploring creativity in AIGC, specifically through human sketches. By enabling a more interactive and dynamic creative process that allows for both text and sketch to inform the ideation process, we hope to unlock new opportunities for creativity in visual content creation. Our method also addresses the limitations of current AIGC approaches and democratises the ideation process by making it accessible to a wider range of users. As the field of AIGC continues to evolve, we believe that exploring the creative potential of human sketches will inspire new directions and advancements that will benefit artists, designers, and creative professionals worldwide.


\bibliography{arxiv}

\end{document}